# A Mixed Graphical Model for Rhythmic Parsing

Christopher Raphael*
Department of Mathematics and Statistics
University of Massachusetts, Amherst
raphael@math.umass.edu

## Abstract

A method is presented for the rhythmic parsing problem: Given a sequence of observed musical note onset times, we simultaneously estimate the corresponding notated rhythm and tempo process. A graphical model is developed that represents the evolution of tempo and rhythm and relates these hidden quantities to an observable performance. The rhythm variables are discrete and the tempo and observation variables are continuous. We show how to compute the globally most likely configuration of the tempo and rhythm variables given an observation of note onset times. Preliminary experiments are presented on a small data set. A generalization to computing MAP estimates for arbitrary conditional Gaussian distributions is outlined.

## 1 Introduction

Music information retrieval (IR) attempts for music what traditional IR does for text. Accordingly, a central challenge of music IR is the generation of music databases in formats amenable to automated search and analysis. While the "right" representation depends on what question one is trying to answer, the most basic elements of music, pitch and rhythm, must almost certainly figure prominently in any usable representation.

Rhythm is the aspect of music that deals with *when* events occur. Typically, rhythm in Western music is notated in a way that expresses the position of each note as a rational number, usually in terms of some relatively small common denominator. For instance, if we use the *measure* as our unit of notated position, then the sequence of measure positions $m_0 = 0, m_1 = 1/4, m_2 = 1, \ldots$ expresses the notion that the first note occurs at the beginning of the 1st measure, the second note occurs 1/4 the way through the 1st measure, the third note occurs at the beginning of the 2nd measure, etc. If the music is performed with mechanical precision then a single number, the *tempo*, will map the measure positions to actual times. For instance, if the tempo is 3 seconds per measure, then the notes would occur at 0 secs, 3/4 secs, 3 secs, etc. However, such a performance would be nearly impossible for the human perform to create, and, moreover, would be undesirable. Much of the expressively quality of a musical performance comes from the way in which the actual note times deviate from what is prescribed by a literal interpretation of the printed music. In particular, there are two primary components to this *expressive timing* [1]. Firstly, the actual tempo is often not constant, but rather continually varied throughout the evolution of the performance. Secondly, there are more local (note by note) distortions which can be accidental, or can result from interpretive considerations.

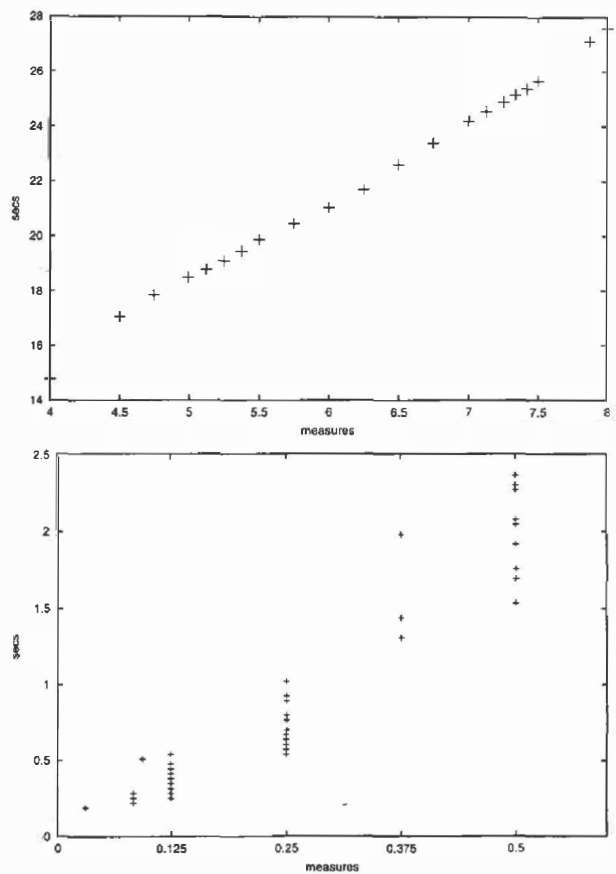

Figure 1: **Top:** Real time (seconds) vs. Musical time (measures) for a musical excerpt. **Bottom:** The actual durations (seconds) of notes grouped by the musical duration (measures).

*This work is supported by NSF grant IIS-9987898.

We focus here on a problem encountered in music IR: Given a sequence of measured note onset times, we wish to identify the corresponding sequence of measure positions. We call this process *rhythmic parsing*. The time sequences forming the input to our procedure could be estimated from an audio file or could come directly from a MIDI (musical instrument digital interface) file — a sequence of time-tagged musical events such as note beginnings and endings. For example, consider the data in the top panel of Figure 1 containing estimated note times from an excerpt of Schumann's 2nd Romance for oboe and piano (oboe part only). The actual audio file can be heard at http://fafner.math.umass.edu/rhythmic_parsing. Our goal is to assign the proper score position, in measures, to each the observed times. When this is done correctly, as in Figure 1, the observed times, in seconds, plotted against the score positions, in measures, trace out a curve whose slope gives the player's time-varying tempo.

We are aware of several several applications of rhythmic parsing. Virtually every commercial score-writing program now offers the option of creating scores by directly entering MIDI data from a keyboard. Such programs must infer the rhythmic content from the actual times at which musical events occur and, hence, must address the rhythmic parsing problem. When the input data is played with anything less than mechanical precision, the transcription degrades rapidly, due to the difficulty in computing the correct rhythmic parse. Rhythmic parsing also has applications in musicology where it could be used to separate the inherently intertwined quantities of notated rhythm and expressive timing. Either the rhythmic data or the timing information could be the focal point of further study. Finally, the musical world eagerly awaits the compilation of music databases containing virtually every kind of (public domain) music. The construction of such data bases will likely involve several transcription efforts including optical music recognition, musical audio signal recognition, and MIDI transcription. Rhythmic parsing is an essential ingredient to the latter two efforts.

Most commercially available programs accomplish this rhythmic parsing task by *quantizing* the observed note lengths, or more precisely inter-onset intervals (IOIs), to their closest note values (eighth note, quarter note, etc.), given a known tempo, or quantizing the observed note onset times to the closest points in a rigid grid [2]. While such quantization schemes can work reasonably well when the music is played with robotic precision (often a metronome is used), they perform poorly when faced with the more expressive and less accurate playing typically encountered. Consider the bottom panel of Figure 1 in which we have plotted the written note lengths in measures versus the actual note lengths (IOIs) in seconds from our musical excerpt. The large degree of overlap between the empirical distributions of each note length class demonstrates the futility of assigning note lengths through note-by-note quantization in this example. In this particular example, the overlap in empirical distributions is mostly attributable to tempo fluctuations in the performance.

We are aware of several research efforts related to rhythm transcription. Some of this research addresses the problem of *beat induction*, or *tempo tracking* in which one tries to estimate a sequence of times corresponding to evenly spaced rhythmic intervals (e.g. beats) for a given sequence of observed note onset times [3], [4]. Another direction addresses the problem of assigning rhythmic values as simple integer ratios to observed note lengths without any corresponding estimation of tempo [1], [5], [6]. The latter two assume that beat induction has already been performed, where as the former assumes that tempo variations are not significant enough to obscure the ratios of neighboring note lengths.

In many kinds of music we believe it will be exceedingly difficult to *independently* estimate tempo and rhythm, as in the previously cited research, since the observed data is formed from a complex interplay between the two. That is, independent estimation of tempo or rhythm leads to a "chicken and egg" problem: One cannot easily estimate rhythm without knowing tempo and vice-versa. In this work we address the problem of *simultaneous* estimation of tempo and rhythm. From a problem domain point of view, this is the most significant contrast between our work and other efforts cited.

The paper is organized as follows. Section 2 develops a generative graphical model for the simultaneous evolution of tempo and rhythm processes that incorporates both prior knowledge concerning the nature of the rhythm process and a simple and reasonable model for tempo evolution. This section then describes a computational scheme for identifying the most likely configuration of the unobserved processes given observed musical data. Section 3 demonstrates the application of our scheme to a short musical excerpt. Section 4 sketches the generalization of our methodology to the generic MAP estimation of unobserved variables for conditional Gaussian (CG) distributions. To our knowledge, MAP estimation in CG distributions has not be studied previously, however is potentially quite useful. Finally, Section 5 lists some easily-derived results about Gaussian kernels that are used in preceding sections.

## 2 Rhythmic Parsing

### 2.1 The Model

Suppose a musical instrument generates a sequence of times $o_0, o_1, \ldots, o_N$, in seconds, at which note onsets occur. Suppose we also have a finite set $\mathcal{S}$ composed of the possible *measure positions* a note can occupy. For instance, if the music is in 4/4 time and we believe that no subdivision occurs beyond the eighth note, then $\mathcal{S} = \{i/8 : i = 0, \ldots, 7\}$. More complicated subdivision rules could lead to sets, $\mathcal{S}$, which are not evenly spaced multiples of some common denominator. We assume only that the possible onset positions of $\mathcal{S}$ are rational numbers in $[0, 1)$, decided upon in advance. Our goal is to associate each note onset $o_n$ with a score position — a measure number and an element of $\mathcal{S}$. For the sake of simplicity we assume that no two of the $\{o_n\}$ can be associated with the exact same score position. This assumption is correct if the times are produced by a monophonic (single voice) instrument and can easily be modified to accommodate polyphonic instruments.

We model this situation as follows. Let $S_0, S_1, \ldots, S_N$ be the discrete measure position process, $S_n \in \mathcal{S}, n = 0, \ldots, N$. In interpreting these positions we assume that each consecutive pair of positions corresponds to a note length of at most one measure. For instance, in the 4/4 example given above $S_n = 0/8, S_{n+1} = 1/8$ would mean the $n$th note begins at the start of the measure and lasts for one eighth note, while $S_n = 0/8, S_{n+1} = 0/8$ would mean the $n$th note begins at

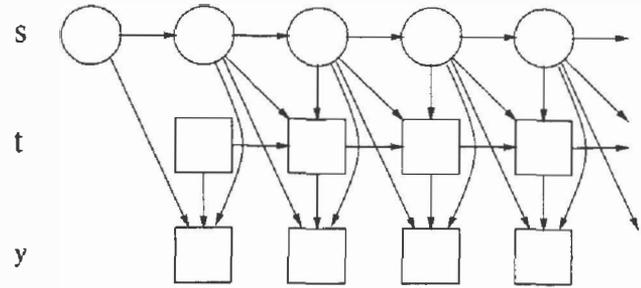

Figure 2: The DAG describing the dependency structure of the variables of our model. Circles represent discrete variables while squares represent continuous variables.

the first eighth note of the measure and lasts for one whole measure. We can then use

$$l(s_n, s_{n+1}) = \begin{cases} s_{n+1} - s_n & \text{if } s_{n+1} > s_n \\ 1 + s_{n+1} - s_n & \text{otherwise} \end{cases}$$

to unambiguously represent the length, in measures, of the transition from $s_n$ to $s_{n+1}$. Thus, if $s_0, s_1, \ldots, s_N$ is known, we assign a score position, $m_n$, to every observation $o_n$ by $m_n = s_0 + \sum_{\nu=1}^{n} l(s_\nu, s_{\nu-1})$. Extending this model to allow for notes longer than a measure complicates our notation slightly, but requires no change of our basic approach. We model the $S$ process as a time-homogeneous Markov chain with initial distribution

$$I(s_0) = P(S_0 = s_0)$$

and transition probability matrix

$$R(s_n, s_{n+1}) = P(S_{n+1} = s_{n+1} | S_n = s_n)$$

The tempo is the most important link between the printed note lengths, $l(S_n, S_{n+1})$, and the observed note lengths, $o_{n+1} - o_n$. Let $T_1, T_2, \ldots, T_N$ be the continuously-valued tempo process, measured in *seconds per measure*, which we model by

$$T_1 \sim N(\nu, \phi^2)$$

and

$$T_n = T_{n-1} + \delta_n$$

for $n = 2, 3, \ldots, N$ where $\delta_n \sim N(0, \tau^2 l(S_{n-1}, S_n))$. This model captures the property that the tempo tends to vary smoothly and that longer notes allow for greater changes in local tempo.

Finally we assume that the observed note lengths $y_n = o_n - o_{n-1}$ for $n = 1, 2, \ldots, N$ are approximated by the product of $l(S_{n-1}, S_n)$ (measures) and $T_n$ (secs. per measure). Specifically

$$Y_n = l(S_{n-1}, S_n)T_n + \epsilon_n$$

where $\epsilon_n \sim N(0, \rho^2 l(S_{n-1}, S_n))$. Note that longer note values are associated with greater differences between the expected and observed note durations. The $\{\delta_n\}$ and $\{\epsilon_n\}$ variables are assumed to be mutually independent.

These modeling assumptions lead to a graphical model whose directed acyclic graph is given in Figure 2. The model is composed of both discrete and Gaussian variables with the property that, for every configuration of discrete variables, the continuous variables have multivariate Gaussian distribution. Thus, the $S_0, \ldots, S_N$, $T_1, \ldots, T_N, Y_1, \ldots, Y_N$ collectively have a conditional Gaussian (CG) distribution.

Such distributions were introduced by Lauritzen and Wermuth [7], [8], and have been developed by Lauritzen [9], and Lauritzen and Jensen [10], in which evidence propagation methodology is described, enabling the computation of local marginal distributions. Using these ideas, we could, in principle, fix $Y_1 = y_1, \ldots, Y_N = y_N$ and proceed to compute marginal distributions on the $\{S_n\}$ and choose as our estimate of $S_n$

$$\bar{s}_n = \arg\max_{s \in S} P(S_n = s | Y_1 = y_1, \ldots, Y_N = y_N)$$

However, these computations rely on construction of a triangulated graph with *strong root*. The additional edges involved in the construction of such a strong root leads to a graph in which a single clique contains the entire collection of $\{S_n\}$ variables. The following computations are intractable. Furthermore, there is no guarantee that the sequence $\bar{s}_0, \ldots, \bar{s}_N$ is reasonable, or even that

$$P(S_0 = \bar{s}_0, \ldots, S_N = \bar{s}_N | Y_1 = y_1, \ldots, Y_N = y_N) > 0$$

strongly calling this estimate into question.

Rather, we desire the *configuration* of unobserved variables which has greatest probability given the observation. Thus, regarding $y_1, \ldots, y_N$ as fixed, we seek the estimate

$$(\hat{s}, \hat{t}) = \arg\max_{s,t} L(s, t, y) \quad (1)$$

where $L(s, t, y)$ is the joint likelihood of $s = (s_0, \ldots, s_N)$, $t = (t_1, \ldots, t_N)$, $y = (y_1, \ldots, y_N)$.

The computation of such MAP estimators for networks composed entirely of discrete variables is well-known [11],[12]. In what follows we demonstrate new methodology for the exact computation of the global maximizer, $(\hat{s}, \hat{t})$ in our mixed discrete and continuous case.

### 2.2 Computing the Rhythmic Parse

Define the $n$-dimensional *Gaussian kernel*

$$K(x; \theta) = K(x; h, m, Q) = h e^{-\frac{1}{2}(x-m)^t Q(x-m)} \quad (2)$$

where $x$ is an $n$-vector, $h$ is a nonnegative constant, $m$ is an $n$-vector, and $Q$ is an $n \times n$ nonnegative definite matrix. We write $\theta = (h(\theta), m(\theta), Q(\theta))$ to represent the components of $\theta$. It is possible to perform a number of operations on Gaussian kernels such as multiplication of two kernels, maximizing over a subset of variables, and representing conditional Gaussian distributions by performing transformations of the parameters involved. Section 5 gives an account of some easily derived results involving Gaussian kernels.

The joint likelihood function $L(s, t, y)$ with $y$ held fixed can be represented as follows. We define

$$\begin{aligned} L_1(s_0, s_1, t_1) &= I(s_0) R(s_0, s_1) N(t_1; \nu, \phi^2) \\ & \quad N(y_1; l(s_0, s_1) t_1, \rho^2 l(s_0, s_1)) \\ &= K(t_1; \theta'(s_0, s_1)) \quad (3) \end{aligned}$$

where $N(\cdot\ ; \mu, \sigma^2) = K(\cdot\ ; (2\pi\sigma^2)^{-1/2}, \mu, 1/\sigma^2)$ is the univariate normal density function. In Eqn. 3, $\theta'(s_0, s_1)$ is computed for each configuration of $s_0, s_1$ by representing the conditional density for $y_1$ as a Gaussian kernel using Eqn. 25, eliminating $y_1$ from the same kernel by holding it fixed using Eqn. 23, multiplying the two kernels together using Eqn. 19, and absorbing the two constants $I(s_0)$ and $R(s_0, s_1)$ into $\theta'(s_0, s_1)$. Using the notation $a_i^j = (a_i, a_{i+1}, \ldots, a_j)$, we then define

$$L_n(s_0^n, t_1^n) = L_{n-1}(s_0^{n-1}, t_1^{n-1}) C(s_{n-1}^n, t_{n-1}^n)$$

for $n = 2, \ldots, N$ where

$$\begin{aligned} C(s_{n-1}^n, t_{n-1}^n) &= R(s_{n-1}, s_n) \\ & \quad N(t_n; t_{n-1}, \tau^2 l(s_{n-1}, s_n)) \\ & \quad N(y_n; l(s_{n-1}, s_n) t_n, \rho^2 l(s_{n-1}, s_n)) \\ &= K(t_{n-1}, t_n; \theta_n^c(s_{n-1}, s_n)) \quad (4) \end{aligned}$$

where Eqn. 4 is computed by representing the two conditional normal densities as Gaussian kernels using Eqn. 25, eliminating $y_n$ from the second density using Eqn. 23, extending the second density to be a function of $t_n$, and $t_{n-1}$ using Eqn. 24, and multiplying the two factors together and absorbing the constant $R(s_{n-1}, s_n)$ using Eqn. 19. Note that $L_N(s_0^N, t_1^N)$ is the joint likelihood $L(s, t, y)$ with $y$ held fixed to the vector of observations. We will compute our MAP estimate by maximizing $L_N$ using dynamic programming as follows.

Define

$$\begin{aligned} H_n(s_n, t_n) &= \max_{s_0^{n-1}, t_1^{n-1}} L_n(s_0^n, t_1^n) \\ & \quad \max_{s_0^{n-1}, t_1^{n-1}} L_n(s_0^{n-1}, t_1^{n-1}, s_n, t_n) \end{aligned}$$

for $n = 1, \ldots, N$. The fundamental observation of dynamic programming is that we can compute $H_n$ recursively by

$$\begin{aligned} H_{n+1}(s_{n+1}, t_{n+1}) &= \max_{s_0^n, t_1^n} L_n(s_0^n, t_1^n) C(s_n^{n+1}, t_n^{n+1}) \\ &= \max_{s_n, t_n} H_n(s_n, t_n) \\ &\quad C(s_n^{n+1}, t_n^{n+1}) \end{aligned} \quad (5)$$

for $n = 1, \ldots, N - 1$.

Consider first the computation of $H_1(s_1, t_1)$ which can be computed by "maxing out" the $s_0$ variable in Eqn. 3. Thus

$$\begin{aligned} H_1(s_1, t_1) &= \max_{s_0} L_1(s_0, s_1, t_1) \\ &= \max_{s_0} K(t_1; \theta'(s_0, s_1)) \\ &= \max_{\theta_1 \in \tilde{\Theta}_1(s_1)} K(t_1; \theta_1) \quad (6) \\ &= \max_{\theta_1 \in \Theta_1(s_1)} K(t_1; \theta_1) \quad (7) \end{aligned}$$

where

$$\tilde{\Theta}_1(s_1) = \{\theta'(s_0, s_1) : s_0 \in \mathcal{S}\}$$

and $\Theta_1(s_1) = \text{Thin}(\tilde{\Theta}(s_1))$ where $\text{Thin}(\Theta)$ is the smallest subset of $\Theta$ such that

$$\max_{\theta \in \text{Thin}(\Theta)} K(t; \theta) = \max_{\theta \in \Theta} K(t; \theta) \quad (8)$$

We remark that it is a simple matter to identify $\Theta_1(s_1) = \text{Thin}(\tilde{\Theta}(s_1))$ since we can "build" the maximum of Eqn. 6 by incrementally adding components of $\tilde{\Theta}_1(s_1)$ while discarding those that leave the maximum unchanged. This algorithm is made more precise in Section 2.3.

The computational feasibility of our dynamic programming algorithm follows because the form of Eqn. 7 — a maximum of Gaussian kernels — is invariant under the operation of Eqn. 5. That is, assuming

$$H_n(s_n, t_n) = \max_{\theta_n \in \Theta_n(s_n)} K(t_n; \theta_n) \quad (9)$$

we have

$$\begin{aligned} H_{n+1}(s_{n+1}, t_{n+1}) &= \max_{s_n, t_n} H_n(s_n, t_n) C(s_n^{n+1}, t_n^{n+1}) \\ &= \max_{s_n, t_n} \max_{\theta_n \in \Theta_n(s_n)} K(t_n; \theta_n) \\ &\quad K(t_n, t_{n+1}; \theta_n^c(s_n, s_{n+1})) \\ &= \max_{s_n, \theta_n \in \Theta_n(s_n)} \max_{t_n} K(t_n; \theta_n)(10) \\ &\quad K(t_n, t_{n+1}; \theta_n^c(s_n, s_{n+1})) \\ &= \max_{s_n, \theta_n \in \Theta_n(s_n)} \quad (11) \\ &\quad K(t_{n+1}; \tilde{\theta}(\theta_n, \theta_n^c(s_n, s_{n+1}))) \\ &= \max_{\theta_{n+1} \in \tilde{\Theta}_{n+1}(s_{n+1})} K(t_{n+1}; \theta_{n+1}) \\ &= \max_{\theta_{n+1} \in \Theta_{n+1}(s_{n+1})} K(t_{n+1}; \theta_{n+1}) \end{aligned}$$

where in going from Eqn. 10 to 11, i. e. in computing $\tilde{\theta}(\theta_n, \theta_n^c(s_n, s_{n+1}))$, we use Eqns. 19, 24, and 22. $\tilde{\Theta}_{n+1}(s_{n+1})$ in the preceding is given by

$$\begin{aligned} \tilde{\Theta}_{n+1}(s_{n+1}) &= \{\tilde{\theta}(\theta_n, \theta_n^c(s_n, s_{n+1})) \\ &\quad : \theta_n \in \Theta_n(s_n), s_n \in \mathcal{S}\} \end{aligned}$$

and $\Theta_{n+1}(s_{n+1}) = \text{Thin}(\tilde{\Theta}_{n+1}(s_{n+1}))$.

While the comparison of Eqns. 9 and 11 suggest $|\Theta_n(s_n)|$ increases exponentially with $n$, this growth will be controlled by the "thinning" operation. In fact, the behavior observed in our experiments, which we anticipate is typical, was that $|\Theta_n(s_n)|$ increased to a manageable number within a few dynamic programming iterations and fluctuated around that number in the following iterations. Details are given in Section 3.

The maximizing value of $L = L_N$ is easily computed as follows. Define

$$\Theta_n(\mathcal{S}) = \bigcup_{s_n \in \mathcal{S}} \Theta_n(s_n) \quad (12)$$

for $n = 1, \ldots, N$ and let

$$(\hat{t}_N, \hat{\theta}_N) = \arg\max_{t_N, \theta \in \Theta_N(\mathcal{S})} K(t_N; \theta)$$

which can be computed by letting

$$\begin{aligned} \hat{\theta}_N &= \arg\max_{\theta \in \Theta_N(\mathcal{S})} (\max_{t_N} K(t_N; \theta)) \\ &= \arg\max_{\theta \in \Theta_N(\mathcal{S})} h(\theta) \end{aligned}$$

and taking $\hat{t}_N = m(\hat{\theta}_N)$. Then

$$\begin{aligned} K(\hat{t}_N; \hat{\theta}_N) &= \max_{t_N} \max_{\theta \in \Theta_N(\mathcal{S})} K(t_N; \theta) \\ &= \max_{s_N, t_N} \max_{\theta \in \Theta_N(s_N)} K(t_N; \theta) \\ &= \max_{s_N, t_N} H_N(s_N, t_N) \\ &= \max_{s_0^N, t_1^N} L_N(s_0^N, t_1^N) \end{aligned}$$

Thus $K(\hat{t}_N; \hat{\theta}_N)$ is the maximal value of the likelihood function, $L$.

We wish to recover the rhythmic parse $\hat{s}_0^N, \hat{t}_1^N$ that attains this maximum. Considering Eqn. 11, we see that each element $\theta_{n+1} \in \Theta_{n+1}(s_{n+1})$ is generated by a unique "predecessor" or "parent" $\text{Pa}(\theta_{n+1}) \in \Theta_n(\mathcal{S})$. That is, if $\theta_n \in \Theta_n(\mathcal{S})$, and

$$\theta_{n+1} = \tilde{\theta}(\theta_n, \theta_n^c(s_n, s_{n+1})) \in \Theta_{n+1}(s_{n+1})$$

then $\text{Pa}(\theta_{n+1}) = \theta_n$. Thus we can trace back the optimizing sequence of parameter values by $\hat{\theta}_n = \text{Pa}(\hat{\theta}_{n+1})$ for $n = 0, \ldots, N - 1$ as in Figure 3. Then the optimizing sequence of measure positions in $\mathcal{S}$ is given by $\hat{s}_n = s(\hat{\theta}_n)$ for $n = 0, \ldots, N$, where $s(\theta_n) = s_n$ if $\theta_n \in \Theta_n(s_n)$.

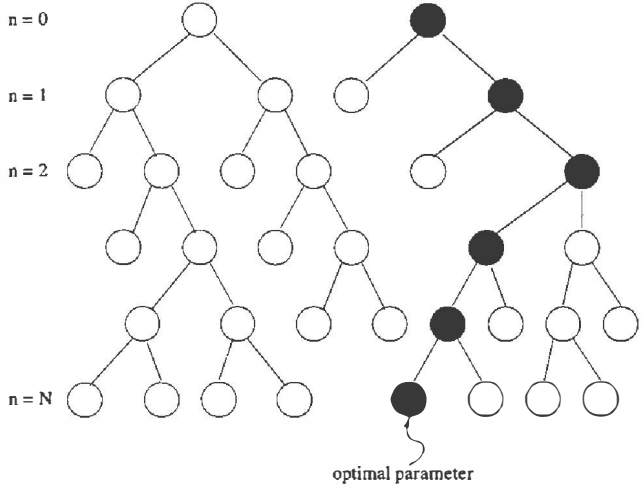

Figure 3: Each parameter $\theta \in \Theta_n(\mathcal{S})$ has a unique parent, $\text{Pa}(\theta)$, so the optimal sequence of parameter values $\theta_0, \ldots, \theta_N$ (shown with solid circles) can be traced back from leaf to root. $\hat{\theta}_N$ is marked as "optimal parameter" in the figure.

Having identified $\hat{t}_N$ and $\hat{s}_0, \ldots, \hat{s}_N$, we can use Eqn. 10 to recover the optimal $\hat{t}_1, \ldots, \hat{t}_N$ through

$$\begin{aligned}
\hat{t}_n &= \arg\max_{\hat{s}_n, t_n} H_n(\hat{s}_n, t_n) C(\hat{s}_n, \hat{s}_{n+1}, t_n, \hat{t}_{n+1}) \\
&= \arg\max_{t_n} K(t_n; \hat{\theta}_n) K(t_n, \hat{t}_{n+1}; \theta_n^c(\hat{s}_n, \hat{s}_{n+1})) \\
&= \arg\max_{t_n} K(t_n; \bar{\theta}_n) \\
&= m(\bar{\theta}_n)
\end{aligned}$$

where $\bar{\theta}_n$ can be computed by eliminating $\hat{t}_{n+1}$ using Eqn. 23 and multiplying the two kernels together using Eqn. 19.

### 2.3 Thinning

The computational feasibility of our dynamic programming algorithm relies on the thinning operation of Section 2.2 since without this operation the complexity of the representation of $H_n$ grows exponentially. Recall, $\text{Thin}(\Theta)$ is the smallest subset of $\Theta$ for which Eqn. 8 holds. When $\Theta$ is composed of parameters for *one-dimensional* Gaussian kernels, as in Section 2.2, the algorithm for computing $\text{Thin}(\Theta)$ is straightforward, as follows.

Suppose $\Theta = \{\theta^1, \ldots, \theta^I\}$. Define

$$\hat{\theta}^i(t) = \arg\max_{\theta \in \{\theta^1, \ldots, \theta^i\}} K(t; \theta)$$

for $i = 1 \ldots I$ and note that $\hat{\theta}^i(t)$ is piecewise constant

and, hence, can be written as

$$\hat{\theta}^i(t) = \sum_{k=1}^{N(i)} \theta_k^i \mathbf{1}_{(x_k^i, x_{k+1}^i)}(t)$$

where $-\infty = x_1^i < x_2^i < \ldots < x_{N(i)}^i < x_{N(i)+1}^i = \infty$ and $\theta_k^i \in \{\theta^1, \ldots, \theta^i\}$. We need not be concerned with the definition of $\hat{\theta}^i(t)$ at points, $t = x_k^i$, where the maximizer is not unique. Clearly then $\text{Thin}(\Theta) = \cup_{k=1}^{N(I)} \theta_k^I$.

Note that $\hat{\theta}^i(t)$ can be computed iteratively by letting $\hat{\theta}^1(t) = \theta^1$ and noting

$$\hat{\theta}^i(t) = \arg\max_{\theta \in \{\hat{\theta}^{i-1}(t), \theta^i\}} K(t; \theta) \quad (13)$$

$\hat{\theta}^i(t)$ can then be computed on each interval $(x_k^{i-1}, x_{k+1}^{i-1})$ where it must be that $\hat{\theta}^{i-1}(t) = \theta_k^{i-1}$. To do this we simply find all solutions, $t$, to the quadratic equation

$$\log K(t; \theta^i) - \log K(t; \theta_k^{i-1}) = 0 \quad (14)$$

which lie in $(x_k^{i-1}, x_{k+1}^{i-1})$. These points partition the interval $(x_k^{i-1}, x_{k+1}^{i-1})$ into subintervals where $\hat{\theta}^i(t)$ must be constant so we need only identify $\hat{\theta}^i(t)$ through Eqn. 13 at any interior point of these subintervals. Having done this for each interval $(x_k^{i-1}, x_{k+1}^{i-1})$ we may find that $\hat{\theta}^i(t)$ is constant over neighboring subintervals. In such a case, the neighbors are simply merged together to form a more compact representation of $\hat{\theta}^i(t)$.

With a minor variation on the thinning algorithm we can, in many cases, compute a *constrained* optimal parse defined by Eqn. 1 subject to $t_{\text{low}} < t_n < t_{\text{high}}$ for $n = 1, \ldots, N$ and fixed constants $t_{\text{low}}$ and $t_{\text{high}}$. This is a helpful restriction in our rhythmic parsing problem since we know that the tempo must always be positive and can reasonably be restricted to be less than some maximum value as well. Such a constraint will also increase the efficiency of our algorithm since it will decrease the number of kernels needed to represent $H_n$ in Eqn. 9.

To find the constrained solution to Eqn. 1 we perform the algorithm presented in Section 2.2 using the thinning algorithm as presented above with the following modification. While computing solutions, $t$, to Eqn. 14, we retain only those that also satisfy $t_{\text{low}} < t < t_{\text{high}}$. A simple argument shows that, if the resulting optimal configuration $(\hat{s}, \hat{t})$ also satisfies $t_{\text{low}} < t_n < t_{\text{high}}$ for $n = 1, \ldots, N$, then $(\hat{s}, \hat{t})$ is the constrained optimal solution.

## 3 Experiments

We performed several experiments using the data depicted in Figure 1 consisting of 129 note times. Since the musical score for this excerpt was available, we extracted the complete set of possible measure positions,

$$S = \left\{ \frac{0}{1}, \frac{1}{8}, \frac{1}{4}, \frac{1}{3}, \frac{3}{8}, \frac{5}{12}, \frac{15}{32}, \frac{1}{2}, \frac{5}{8}, \frac{3}{4}, \frac{7}{8} \right\}$$

The most crucial parameters in our model are those that compose the transition probability matrix $R$. The two most extreme choices for $R$ are the uniform transition probability matrix

$$R^{\text{unif}}(s_i, s_j) = 1/|S|$$

and the matrix ideally suited to our particular recognition experiment

$$R^{\text{ideal}}(s_i, s_j) = \frac{|\{n : S_n = s_i, S_{n+1} = s_j\}|}{|\{n : S_n = s_n\}|}$$

$R^{\text{ideal}}$ is unrealistically favorable to our experiments since this choice of $R$ is optimal for recognition purposes and incorporates information normally unavailable; $R^{\text{unif}}$ is unrealistically pessimistic in employing no prior information whatsoever. The actual transition probability matrices used in our experiments were convex combinations of these two extremes

$$R = \alpha R^{\text{ideal}} + (1 - \alpha) R^{\text{unif}}$$

for various constants $0 < \alpha < 1$. A more intuitive description of the effect of a particular $\alpha$ value is the *perplexity* of the matrix it produces: $\text{Perp}(R) = 2^{H(R)}$ where $H(R)$ is the $\log_2$ entropy of the corresponding Markov chain. Roughly speaking, if a transition probability matrix has perplexity $M$, the corresponding Markov chain has the same amount of "indeterminacy" as one that chooses randomly from $M$ equally likely possible successors for each state. The extreme transition probability matrices have

$$\text{Perp}(R^{\text{ideal}}) = 1.92$$
$$\text{Perp}(R^{\text{unif}}) = 11 = |S|$$

In all experiments we chose our initial distribution, $I(s_0)$, to be uniform, thereby assuming that all starting measure positions are equally likely. The remaining constants, $\nu, \phi^2, \tau^2, \rho^2$ were chosen to be values that seemed "reasonable."

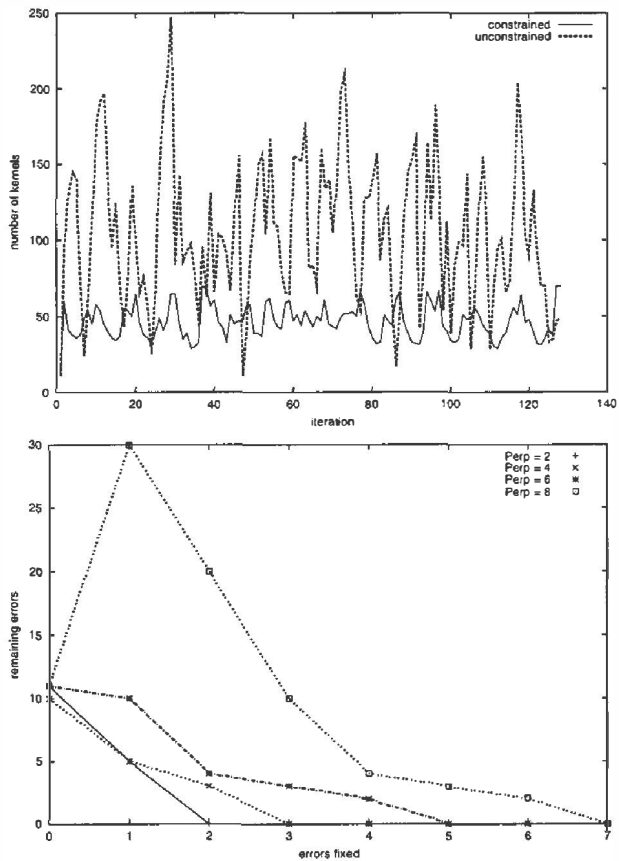

Figure 4: **Top:** The number of Gaussian kernels necessary to represent $H_n$, $|\Theta_n(S)|$, as a function of $n$. **Bottom:** The number of errors produced by our system at different perplexities and with different numbers of errors already corrected.

The computational feasibility of our approach relies on the representation of $H_n$ from Eqn. 9 staying manageably small as $n$ increases. The top panel of Figure 4 shows the evolution of $|\Theta_n(S)|$ for $n = 0, \ldots, 128$ with Perp($R$) = 4. The figure shows results for both the basic algorithm presented in Section 2.2 and for the constrained version discussed in Section 2.3. In the latter version we constrained the tempo variables to lie in $(1,5)$ corresponding to a range of 48 to 240 beats per minute (the composer's tempo marking was 104 beats per minute). Both versions show that the complexity of the representation of $H_n$ does not grow as $n$ increases. The average number of kernels used in the representation of $H_n(s_n, t_n)$, $\sum_{n=0}^{128} |\Theta_n(S)|/(129 \times |S|)$, was 4.22 in the constrained case and 9.59 in the unconstrained case.

The rhythmic parsing problem we pose here is based solely on timing information. Even with the aid of pitch and interpretive nuance, trained musicians occasionally have difficulty parsing rhythms. For this reason, it is not terribly surprising that our parses contained errors. However, a virtue of our approach is that the parses can be incrementally improved by allowing the user to correct individual errors. These corrections are treated as constrained variables in subsequent passes through the recognition algorithm. Such a technique may well be useful in a more sophisticated music recognition system in which it is unrealistic to hope to achieve the necessary degree of accuracy without the aid of a human guide. In Figure 4 we show the number of errors produced under various experimental conditions. The four traces in the plot correspond to perplexities $2, 4, 6, 8$, while each individual trace gives the number of errors produced by the recognition after correcting $0, \ldots, 7$ errors. In each pass the first error found from the previous pass was corrected. All of these experiments were performed with the $\{t_n\}$ constrained to line in $(1,5)$ as discussed in Section 2.3. Two of the $129 \times 8 \times 4 = 4128$ estimated tempo variables were slightly outside this range. Thus, all but two of our parses are exact constrained MAP estimates, while the other two are likely very good approximations.

## 4 Generalization

While the methodology in Section 2 was developed for the particular graphical model of Figure 2, the ideas extend to arbitrary graphical models for conditional Gaussian distributions. While a complete description of this generalization is beyond the scope of this paper, we sketch here such an extension. A complete description of this work can be found in [13]

A mixed collection of discrete and continuous variables, $X$, has a conditional Gaussian (CG) distribution if, for every configuration of the discrete variables, the conditional distribution on the continuous variables is multivariate Gaussian [7], [8]. We assume we have a representation of the CG distribution in terms of a DAG in which discrete nodes have no continuous parents.

If some of the components of $X$ are observed, we can factor the conditional density on the remaining unobserved variables, $X_U$, as

$$\tilde{f}(x_U) = \prod_{C \in \mathcal{C}} \phi_C(x_C) \qquad (15)$$

where $\mathcal{C}$ are the cliques of a junction tree and the potential functions, $\phi_C(x_C)$, depend only on the indicated variables. When $C$ contains continuous variables, $\phi_C$ can be shown to be of the form

$$\phi_C(x_C) = K(x_{\Gamma(C)}; \theta(x_{\Delta(C)})) \qquad (16)$$

where $\Gamma(C)$ and $\Delta(C)$ index the continuous and discrete variables of $C$. Otherwise $\phi_C$ is the usual discrete potential.

The idea of dynamic programming can be extended beyond the linear graph structure encountered in Section 2, to maximize a function of the form of Eqn. 15 with clique potentials as in Eqn. 16. In this context, we define

$$H_C(x_C) = \max_{x_{U \setminus C}} \prod_{\tilde{C} \leq C} \phi_{\tilde{C}}(x_{\tilde{C}})$$

where $\tilde{C} < C$ if $C$ lies on the unique path between $\tilde{C}$ and the root of the junction tree. Then $H_C$ can be computed recursively by the dynamic programming iteration

$$H_C(x_C) = \phi_C(x_C) \prod_{C \xrightarrow{S} \tilde{C}} \max_{x_{\tilde{C} \setminus S}} H_{\tilde{C}}(x_{\tilde{C}}) \qquad (17)$$

where we take $C \xrightarrow{S} \tilde{C}$ to mean that $C$ and $\tilde{C}$ are neighboring cliques separated by $S$ with $\tilde{C} < C$.

As in Section 2, a specific functional form can be used to represent the $H_C$ functions throughout the dynamic programming recursion. Suppose $C$ has continuous components and consider the form

$$H_C(x_C) = \max_{\theta \in \Theta(x_{\Delta(C)})} K(x_{\Gamma(C)}; \theta) \qquad (18)$$

($H_C$ is just a non-negative function when $x_C$ has only discrete components). The terminal cliques clearly have $H_C$ of this form, where the maximum has a single Gaussian kernel. Furthermore one can show that if all child cliques of $C$ have such a representation, then the Eqn. 17 also leads to a similar representation for $H_C$. Having computed the $H_{C_r}$ for the root clique, $C_r$ we can easily trace back the calculations to find the optimal configuration $\hat{x}_U$.

While most of the methodology presented in Section 2 extends in a straightforward manner to the general domain of CG distributions, there is one notable exception. The computation of Eqn. 18 also involves the thinning operation of Eqn. 8, however, the collection of kernels we consider are not necessarily one-dimensional. Thus, the algorithm of Section 2.3, which is inherently one-dimensional, cannot be applied. We do anticipate that a smarter algorithm can be used to compute, or at least approximate, the thinning operation in higher dimensions. The development of such an algorithm is the only missing link between our proposed methodology and a fully general approach to finding MAP estimates for unobserved variables in CG distributions.

## 5 Gaussian Kernels

A Gaussian kernel is a multivariate function of the form of Eqn. 2. The following identities hold for such functions. The derivations of these results are quite straightforward and are not included here.

**Multiplication:**

$$K(x;h_1,m_1,Q_1)K(x;h_2,m_2,Q_2) = K(x;h,m,Q) \quad (19)$$

where

$$\begin{aligned} h &= h_1 h_2 e^{-\frac{1}{2}(m_1^t Q_1 m_1 + m_2^t Q_2 m_2 - m^t Q m)} \\ m &= Q^{-}(Q_1 m_1 + Q_2 m_2) \\ Q &= Q_1 + Q_2 \end{aligned}$$

where $Q^-$ is the generalized inverse of $Q$.

**Maxing Out:** Let $m$ and $Q$ be partitioned as

$$m = \begin{pmatrix} m_1 \\ m_2 \end{pmatrix} \quad (20)$$

$$Q = \begin{pmatrix} Q_{11} & Q_{12} \\ Q_{21} & Q_{22} \end{pmatrix} \quad (21)$$

Then

$$\max_{x_2} K(\begin{pmatrix} x_1 \\ x_2 \end{pmatrix}; h, m, Q) = K(x_1; h, m_1, \tilde{Q}) \quad (22)$$

where

$$\tilde{Q} = Q_{11} - Q_{12} Q_{22}^{-} Q_{21}$$

**Fixing Variables:** Let $m$ and $Q$ be partitioned as in Eqns. 20,21. Regarding $x_2$ as fixed

$$K(\begin{pmatrix} x_1 \\ \tilde{x}_2 \end{pmatrix}; h, m, Q) = K(x_1; \tilde{h}, \tilde{m}, \tilde{Q}) \quad (23)$$

where

$$\begin{aligned} \tilde{h} &= h e^{-\frac{1}{2}(x_2 - m_2)^t (Q_{22} - Q_{21} Q_{11}^{-} Q_{12})(x_2 - m_2)} \\ \tilde{m} &= m_1 - Q_{11}^{-} Q_{12}(x_2 - m_2) \\ \tilde{Q} &= Q_{11} \end{aligned}$$

**Extension:** The kernel $K(x_1; h, m, Q)$ can be viewed as a function of $x_1$ and $x_2$ by

$$K(x_1; h, m, Q) = K(\begin{pmatrix} x_1 \\ x_2 \end{pmatrix}; h, \begin{pmatrix} m \\ 0 \end{pmatrix}, \begin{pmatrix} Q & 0 \\ 0 & 0 \end{pmatrix}) \quad (24)$$

**Conditional Gaussian Densities:** If $x_2 = \alpha^t x_1 + \beta + \xi$ where $x_2$ is univariate and $\xi \sim N(\mu, \sigma^2)$, the conditional density of $x_2$ given $x_1$ is

$$f_v(x_2|x_1) = K(\begin{pmatrix} x_1 \\ x_2 \end{pmatrix}; h, m, Q) \quad (25)$$

where

$$\begin{aligned} h &= (2\pi\sigma^2)^{-1/2} \\ m &= \begin{pmatrix} 0 \\ \beta + \mu \end{pmatrix} \\ Q &= \frac{1}{\sigma^2} \begin{pmatrix} \alpha\alpha^t & -\alpha \\ -\alpha^t & 1 \end{pmatrix} \end{aligned}$$